\documentclass[runningheads]{llncs}
\usepackage[T1]{fontenc}
\usepackage{graphicx}
\usepackage{cite}
\usepackage{amsmath,amssymb,amsfonts}
\usepackage{algorithmic}
\usepackage{graphicx}
\usepackage{textcomp}
\usepackage{xcolor}

\usepackage{subcaption}
\usepackage{todonotes}

\usepackage{tikz}
\newcommand*{\myhash}{%
  \begin{tikzpicture}
    \pgfmathsetlengthmacro\myWidth{.8*width("=")}%
    \pgfmathsetlengthmacro\myHeight{height("H")}%
    \pgfmathsetlengthmacro\mySepY{.3333*\myWidth}%
    \pgfmathsetlengthmacro\mySideBearing{.1*\myWidth}%
    \def\myAngle{70}%
    \pgfmathsetlengthmacro\mySepX{\mySepY/sin(\myAngle)}%
    \pgfmathsetlengthmacro\mySlantX{\myHeight/tan(\myAngle)}%
    \draw[line cap=round]
      (0, {(\myHeight - \mySepY)/2}) -- ++(\myWidth, 0)
      (0, {(\myHeight + \mySepY)/2}) -- ++(\myWidth, 0)
      ({(\myWidth - \mySepX - \mySlantX)/2}, 0)
      -- ({(\myWidth - \mySepX + \mySlantX)/2}, \myHeight)
      ({(\myWidth + \mySepX - \mySlantX)/2}, 0)
      -- ({(\myWidth + \mySepX + \mySlantX)/2}, \myHeight)
    ;%
    \useasboundingbox
      (-\mySideBearing, 0)
      (\myWidth + \mySideBearing, \myHeight)
    ;% 
  \end{tikzpicture}%
}

\newcommand{\uc}{\mathcal{C}}
\newcommand{\ua}{\mathcal{A}}
\newcommand{\ut}{\mathcal{T}}
\newcommand{\ud}{\mathcal{D}}
\newcommand{\ue}{\mathcal{E}}
\newcommand{\tf}{\mathit{tf}}
\newcommand{\utf}{\mathit{\mathcal{TF}}}
\newcommand{\us}{\mathcal{S}}

\newcommand{\ul}{\mathcal{L}}
\newcommand\prfx{\mathit{prfx}}
\newcommand\surp{\mathit{surp}}
\newcommand\siml{\mathit{sim}}
\newcommand\eff{\mathit{eff}}
\newcommand\avg{\mathit{avg}}
\newcommand{\sit}{\mathit{sit}}

\newcommand{\uv}{\mathcal{V}}
\newcommand{\cov}{\mathit{cov}}

\begin{document}
\title{Detecting Surprising Situations in Event Data
}
\author{Christian Kohlschmidt \and Mahnaz Sadat Qafari \and
Wil M. P. van der Aalst}
\authorrunning{C. Kohlschmidt et al.}
\institute{Process and Data Science Chair (PADS)\\
\textit{RWTH Aachen University}
Aachen, Germany \\
\email{christian.kohlschmidt@rwth-aachen.de, \{m.s.qafari,wvdaalst\}@pads.rwth-aachen.de}}
\maketitle

\begin{abstract}
Process mining is a set of techniques that are used by organizations to understand and improve their operational processes. The first essential step in designing any process reengineering procedure is to find process improvement opportunities. In existing work, it is usually assumed that the set of problematic process instances in which an undesirable outcome occurs is known prior or is easily detectable. So the process enhancement procedure involves finding the root causes and the treatments for the problem in those process instances. For example, the set of problematic instances is considered as those with outlier values or with values smaller/bigger than a given threshold in one of the process features. However, on various occasions, using this approach, many process enhancement opportunities, not captured by these problematic process instances, are missed. To overcome this issue, we formulate finding the process enhancement areas as a context-sensitive anomaly/outlier detection problem. We define a process enhancement area as a set of situations (process instances or prefixes of process instances) where the process performance is surprising. We aim to characterize those situations where process performance/outcome is significantly different from what was expected considering its performance/outcome in similar situations. To evaluate the validity and relevance of the proposed approach, we have implemented and evaluated it on several real-life event logs.
\end{abstract}

\section{Introduction}
Considering the current highly competitive nature of the economy, it is vital for organizations to continuously improve/enhance their processes in order to meet the best market standards and improve customer experience. Process enhancement involves many steps, including finding the process areas where improvements are possible, designing the process reengineering steps, and estimating the impact of changing each factor on the process performance. By conducting all these steps, organizations can benefit from applying process mining techniques. 
The first step of process enhancement is detecting those process areas where an improvement is possible. Process mining includes several techniques for process monitoring and finding their friction points. However, these techniques have the hidden assumption that all the process instances (cases) are the same. So the set of problematic cases can be easily identified. For example, the problematic cases can be identified as the ones with an outlier value with respect to a process feature. Another common method is using a threshold for a specific process feature. However, considering the variety of the cases, it is possible that a solution solves the problem for one group of cases while aggravating the problem for another group. Moreover, using the current techniques, the performance of the process in some cases can be considered normal/acceptable compared to the overall behavior of the process, while it can be considered surprising (anomalous/undesirable) when just considering their similar cases. This phenomenon can lead to overlooking some of the process enhancement opportunities. 

As another issue, there are several process instances where the process performs significantly better than other similar process instances. Analyzing the process behavior while performing these process instances can lead to invaluable clues on how to improve the process. Usually, this source of information is neglected by the current process mining techniques.

\begin{figure*}[b]
        \centering
            \includegraphics[width = 125mm]{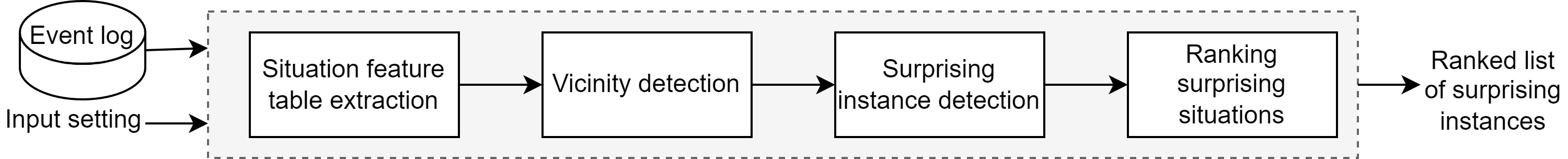}
            \caption{The general overview of the proposed method.}
            \label{fig::generalView}
    \end{figure*}

To overcome these issues, we formulate finding those areas where a process enhancement is possible, as the problem of finding those groups of process situations where the process performance is significantly different from their similar situations. Here, we define a \emph{process situation} (or simply a \emph{situation}) as a process instance, or a prefix of it. The proposed method includes four steps, (1) enriching and extracting the data from the event log, (2) finding a set of sets of similar situations (which we call a \emph{vicinity cover} and each set of similar situations is a \emph{vicinity}). Naturally, a measure is needed to measure the similarity between instances and identify vicinities. However, having access to such a measure is a strong assumption. Thus we use a machine learning technique to determine the vicinities in the absence of such a measure. (3) The next step involves finding the set of surprising situations in each vicinity (if any exist). (4) Finally, a list of detected sets of surprising situations is presented to the user ordered by their effect on the process and how surprising they are. These findings can be further analyzed to understand the reason for the different behavior of the process in these surprising situations and gain insights on how to improve the process. 
Figure \ref{fig::generalView} shows the general overview of the proposed method. 

\begin{figure}[t]
        \centering
            \includegraphics[width=60mm]{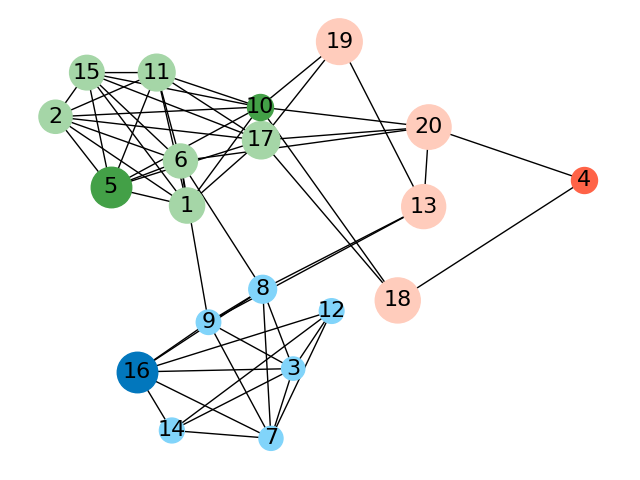}
            \caption{A graph representing the similarity of situations in a loan application example. Each node represents a situation (a process instance). Two situations are similar if the Levenshtein distance of their activity sequences is at most one. The vicinity of a node is the set of process instances in the same community. Three vicinities have been detected in this example, which are colored red, blue, and green. Surprising situations are highlighted with a darker color. The throughput of each situation is proportional to the size of its corresponding node.}
            \label{fig::example}
    \end{figure}

For example, consider that in a loan application process with 20 cases, we are interested in finding those cases where their throughput is surprising. In this example, each process instance (case) is a situation. Also, we consider two situations similar if the Levenshtein distance of their activity sequence is at most one. Figure \ref{fig::example} shows the graph for the cases of this loan application, where each case corresponds to a node. Two cases are connected if they are similar. The size of each node is proportional to its throughput. The colors (blue, green, and red) indicate the vicinities found by the Louvain community detection algorithm. The nodes highlighted with darker colors are the surprising cases where the throughput time is significantly different from the other cases in the same vicinity. In this example, the throughput was worse than expected for cases 5 and 16 and better than expected for cases 4 and 10. The process owner can gain actionable insights by analyzing the behavior of the process in these cases, particularly in comparison with their vicinity, to improve/enhance the performance of the process in other similar cases in the future. Note, if we just had considered the overall performance of this process, these four cases would not have been detected. 

The rest of the paper is organized as follows. In Section \ref{sec::relatedWork}, a brief overview of the related work is given. %In Section \ref{sec::preliminaries}, we present preliminaries briefly.
In Section \ref{sec::method}, the proposed method is presented. The experimental results are discussed in Section \ref{sec::results}. Finally, in Section \ref{sec::conclusion}, the conclusion is presented.

\section{Related work}\label{sec::relatedWork}
Existing research on context-aware anomaly detection in process mining is closest to our work. Here we provide an overview of anomaly detection techniques. 
%14bezerra2009anomaly,15bezerra2008anomaly,

Most existing methods investigate anomalies considering the control-flow perspective (e.g., \cite{10van2012replaying,11dees2017enhancing,nguyen2016deviancemining,13pauwels2019anomaly,16bezerra2011fraud,17bezerra2012dynamic}). These methods generate a reference model from the event log and apply conformance checking to detect anomalous behavior. A subgroup of these methods known as \emph{deviance mining approaches} investigate performance anomalies. \cite{nguyen2016deviancemining}. In \cite{10van2012replaying}, the authors identify deviations and bottlenecks by replaying the event log on an enrich process model with performance information. In \cite{11dees2017enhancing}, the authors analyze the deviations between a process model and an event log to identify which deviations enforce positive performance. In \cite{18gupta2015pariket}, the anomalous traces in event logs are detected using window-based and Markovian-based techniques. The drawback of control-flow approaches is that they ignore a wide range of non-control-flow data, which can be used for more sophisticated context-sensitive anomaly detection methods.

%Data-aware anomaly detection combines the process mining and data mining perspective. Approaches in this category integrate process-related data. For example, 
The authors of \cite{19bohmer2016multi} propose an anomaly detection approach that incorporates perspectives beyond the control-flow perspective, such as time and resource-related information. This approach marks events as anomalies based on a certain likelihood of occurrence, however, case anomalies are not considered.

Other approaches in this category only focus on specific use cases. The authors of \cite{20rieke2013fraud} analyze suspicious payment transactions to identify money laundering within a money transfer service. They propose an approach to match the transactions with the expected behavior given by a process model to identify many small transactions that end up on the same account. \cite{21rogge2014temporal} identifies surprisingly short activity execution times in a process by automatically inferring a Bayesian model from the Petri net representation of the process model. The authors of \cite{22sarno2020anomaly} use fuzzy association rule learning to detect anomalies. %They introduce six patterns of fraud, including skipped activities, wrong throughput time, or wrong resources. 
As these approaches specialize in specific use cases, they do not apply to identify anomalies in a general process.

A third category is domain-based anomaly detection. For example, the authors of \cite{23post2021active} propose an approach that supports the identification of unusual or unexpected transactions by encoding the cases and assigning an anomaly score to each case. They use the domain knowledge of domain experts to update the assigned anomaly scores. The approaches in this category require domain knowledge to label cases, which limits their applicability.

\section{Method}\label{sec::method}
Process mining techniques usually start by analyzing an event log. An event log is a collection of cases where each case is a sequence of events, in which each event refers to a case, an activity, and a point in time. More formally, 
    \begin{definition}[Event, Case, Event log]
        Let $\uc$ be the universe of case identifiers, $\ua$ be the universe of activities, $\ut$ be the universe of timestamps. Moreover, let $\ud = \{\ud_1, \dots ,\ud_n\}$ be the universe of domain-dependent data attributes. We define the universe of events as $\ue = \uc \times \ua \times \ut \times \ud_1 \times \dots \times \ud_N$ and each element $e = (c, a, t, d_1, \dots , d_n) \in \ue$ an \emph{event}.
        Let $\ue^+$ be the universe of (non-empty) finite and chronologically ordered sequences of events. We define a \emph{case} as a sequence of events $\sigma \in \ue^+$ in which all events have the same case identifier; i.e. $\forall e_i, e_j \in \sigma  \pi_c(e_i) = \pi_c(e_j)$ where $\pi_c (e)$ returns the case identifier of event $e \in \ue$. We define an \emph{event log}, $L$, as a set of cases in which each case has a unique case identifier; i.e., $\forall \sigma, \sigma' \in L (\exists e \in \sigma \exists e' \in \sigma  \pi_c(e) = \pi_c(e')) \implies \sigma =\sigma' $. We denote the universe of all event logs with $\ul$.
    \end{definition}

We assume that we know the process feature that captures the property of the process that the process owner is interested in its optimization. We call this feature \emph{target feature} and denote it with $\tf$ where $\tf \in \utf = \ua \times \ud$. Note that the target is composed of an attribute name and an activity name, which indicate the attribute value should be extracted from the events with that activity name. The attribute name can be any of the attributes captured by the event log or a derived one. Moreover, we assume that we know \emph{descriptive features}, which are the set of process features that are relevant in measuring the similarity of the situations. In the following, we explain the surprising situation detection steps.
\subsection{Situation Feature Table Extraction}

To find the surprising situations, we have to extract the data in the form of tabular data from the event log. As the detected surprising situations are meant to be used for root cause analysis, to respect the temporal precedence of cause and effect, indicating that the cause must occur before the effect, we extract the data from that prefix of the case that has been recorded before the target feature. We call such a prefix of a case a \emph{situation}. More formally:
\begin{definition}[Situation]\label{def::situation}
    Let $L \in \ul$, $\sigma = \langle e_1, \dots , e_n \rangle \in L$, and \\$\prfx(\langle e_1,\dots , e_n \rangle ) = \{\langle e_1,\dots , e_i \rangle \mid 1\leq i \leq n \} $, a function that returns the set of non-empty prefixes of a given case.
    We define the universe of all situations as $\us= \bigcup_{L\in\ul} S_L$ where $S_L = \{\sigma \mid \sigma \in \prfx (\sigma ') \wedge \sigma' \in L \}$ is the set of situations of event log $L$. We call each element $\sigma \in \us$ a \emph{situation}. Moreover, we define $\sit \in  (\ul \times \utf) \times 2^\us$ to be the a function that returns $\{ \sigma \in S_L \mid \pi_{a} (\sigma) = \mathit{act}\}$ for a given $L\in \ul$ and $\tf = (\mathit{att}, \mathit{act}) \in \utf$, where $\pi_{a}(\sigma)$ returns the activity name of the last event of $\sigma$.
\end{definition}

We call the data table created by extracting data from situations a \emph{situation feature table}. Please note that each row of the situation feature extracted from $\sit(L, \tf)$ corresponds to a situation in it and this correspondence forms a bijection. To enrich the event log and extract the situation feature table, we use the method presented in \cite{qafari2022feature}.

\subsection{Vicinity detection}
Informally, a vicinity is a set of similar situations and a vicinity cover of $S\subseteq \us$ is a set of vicinities of its situations such that their union covers $S$. Let $\cov \in 2^\us \to 2^{2^\us}$ in which $\forall S \subseteq \us \forall S' \in \cov (S) \big( S' \neq \emptyset \wedge (\forall \sigma, \sigma' \in S'  \siml(\sigma, \sigma')=1 )\big)$ and $\forall S \subseteq \us \cup_{S' \in \cov(S)}S' = S$. Here, $\siml \in \us \times \us \to \{0,1\}$ is an indicator function indicating if $\sigma$ and $\sigma'$ are similar, for $\sigma, \sigma' \in \us$.

Using a coverage function, we define a vicinity cover of a set of situations extracted from an event log with respect to a specific target feature as follows:

\begin{definition}[Vicinity and Vicinity Cover]\label{def::cov}
    Let $S = \sit(L, tf)$ be the set of situations extracted from $L \in \ul$ with respect to the target feature $\tf \in \utf$ and $\cov \in 2^\us \to 2^{2^\us}$ be a coverage function. We simply define a \emph{vicinity cover} of $S$ as $\cov(S)$ and we call each member of $V \in \cov(S)$ a \emph{vicinity} of $S$. We denote the universe of all vicinities by $\uv$.
\end{definition}

In the sequel, we explain the vicinity detection method separately for the case where we know the similarity measure and the case where such a similarity measure is not known.

\subsubsection{Vicinity Detection With a Similarity Measure.}
Let $d \in \us \times \us \to \mathbb{R}$ be a distance measure. Then we can say a situation is similar to another situation if their distance is less than $\alpha$. Now, we can define the similarity function as $\siml_{d,\alpha} \in \us \times \us \to \{0, 1\}$ such that $\siml_{d, \alpha} (\sigma_1, \sigma_1)$ returns 1 if $d(\sigma, \sigma') \leq \alpha$ and 0 o.w. for all $\sigma, \sigma' \in \us$. In this case, we can determine the vicinity cover of the set of situations through the coverage function (Definition \ref{def::cov}) in which $\siml_{d,\alpha}(.,.)$ is the similarity function. Another method is to create a graph $G=(S, E)$ in which each node corresponds to one of the situations extracted from the event log. There is an edge between two nodes if the distance of their corresponding situations is smaller than $\alpha$. Using a \emph{community detection} algorithm on this graph, we can determine the vicinities. Note that in this case two situations are similar if their corresponding nodes are in the same community and each detected community is a vicinity. A community detection function aims at finding (potentially overlapping) sets of nodes that optimize the modularity within the similarity graph. Modularity measures the relative density of edges inside the communities compared to edges outside the communities.

\subsubsection{Vicinity Detection Without a Similarity Measure.}
The availability of a distance function is a strong assumption. Considering the complexity of the real-life event data, even for specialists, it is a challenging task to determine such a distance function. Hence, we use machine learning techniques to detect surprising situations in the data. In this case, the process expert needs to know the set of process features relevant to measuring the similarity of the situations and not the exact distance measure. Here we briefly mention the vicinity detection method using a clustering and classification model. 

\paragraph{Surprising Situation Detection Using a Clustering Model} 

We use $k$-means as the clustering model to explain the method; however, the general idea is similar to using other clustering models. To find the surprising situations using a clustering model, we first cluster the situations using $k$-means, with a predefined $k$, based on their descriptive features. In this method, two situations are similar if they belong to the same cluster and each cluster forms a vicinity.% Surprising situations are detected using the same way as in Definition \ref{def::ss}.

\paragraph{Surprising Situation Detection Using a Classification Model} 
We mainly use a decision tree as the classification model. We train a decision tree on the data trying to predict the target feature $\tf$ using descriptive features. In this method, we consider two situations similar if they belong to the same node of the tree. Moreover, we consider the set of situations corresponding to each node of the decision tree (or each node in a subset of nodes of the decision tree, such as leaves) as a vicinity.  %We define the vicinity of each situation $s$ as the set of the situations in the same (internal or leaf) node as $s$. Please note that in this method, each situation may participate in more than one vicinity. %, and we consider the surprising situations as the ones that are miss-classified.

\subsection{Surprising Situation Detection}
We define the surprising situations in each vicinity as those situations in that vicinity that significantly differ from the other situations (in that vicinity). Suppose that $D \in \uv \to \bigcup_{V \in \uv}2^V$ where $\forall V \in \uv : D(V) \subseteq V$ is a function that, given a set of similar situations (a vicinity), returns its subset of surprising ones. We call such a function a \emph{detector}.  For example, a detector function can be a function that returns the subset of situations that exceed a user-defined threshold value for the target feature. Using this function, we define the set of surprising situations of a vicinity as follows: %However, to maximize the applicability of the method and the minimize the needed domain knowledge, we use the boxplot method to find the surprising situations in each vicinity. More precisely,

\begin{definition}[Surprising Situation Set]\label{def::ss}
    %Let $S$ be the set of situations, $V \subseteq S$ a vicinity, and $\gamma \in \mathbb{R}^+$ the surprisingness threshold. We define a situation $s \in V$ surprising if $\mid \pi_\tf (s) - \frac{\sum_{s' \in V)} \pi_\tf (s')}{\mid V\mid } \mid \geq \gamma$ where $\pi_\tf (s)$ returns the value of $\tf$ for $s \in S$.
    %Let $S \in \us$ be the set of situations, $V \subseteq S$ a vicinity, and $\tf$ the target feature. We define a situation $s \in V$ surprising if $s$ is detected as outlier by applying boxplot on the situations in $V$ with regards to $\tf$.
    Let $V \in \uv$ be a vicinity and  $D \in \uv \to \bigcup_{V \in \uv}2^V$ where $\forall V \in \uv : D(V) \subseteq V$ be a detector function. We define $D(V)$ as the set of \emph{surprising situations} in $V$.
\end{definition}
We can find the set of all sets of surprising situations of the set of situations by applying the detector function on all the vicinities of its vicinity cover.
\begin{definition}[Surprising Situation Sets]\label{DEf:sss}
    Let $S = \sit(L, tf)$ be the set of situations extracted from $L \in \ul$ with respect to target feature $\tf \in \utf$, $\cov(S)$ a vicinity cover of $S$, and detection function $D \in \uv \to \bigcup_{V \in \uv}2^V$. We define the \emph{surprising situation sets} of $S$ as $\{D(V)\mid V \in \cov(S) \}$.
\end{definition}

\subsection{Ordering Surprising Situations}
We define two criteria to order the detected surprising situations: \emph{surprisingness} and \emph{effectiveness}. Suppose $U$ is the set of surprising situations in a vicinity $V$. Surprisingness of $U$ measures how rare it is to see such a situation in its vicinity, whereas effectiveness measures how beneficial it is to enhance the process based on the findings of root cause analysis of $U$. More precisely:
\begin{definition}
    Let $V \in \uv$ be a vicinity and $U \subseteq V$ the set of surprising situations in $V$, and $\gamma \in (0,1]$ a threshold. We define the \emph{surprisingness} of $U$ as:
    \begin{equation*}
        \surp(U) = \gamma \mid \avg(U) - \avg (V\setminus U) \mid + (1-\gamma) \frac{\myhash ( U ) }{\myhash ( V )}
    \end{equation*}
    and the \emph{effectiveness} of $U$ as:
    \begin{equation*}
    \eff(U) =
        \begin{cases}
            (\avg(V \setminus U)-\avg(U)) \times \myhash ( V \setminus U)  & \avg(U) < \avg(V \setminus U)\\
            (\avg(U)-\avg(V \setminus U)) \times \myhash ( U ) & \avg(U) > \avg(V \setminus U)
    \end{cases}
    \end{equation*}
    where $\myhash (A)$ denotes the cardinality of $A$ and $\avg(A)= \frac{\sum_{s \in A} \pi_\tf(s)}{\myhash (A)}$ for each $A \subseteq \us$ is the average value of the target feature $tf$ for the situations in $A$.    
\end{definition}
In the above definition, we assume that the lower values for $\tf$ are more desirable. If this assumption does not hold, the effectiveness can be similarly defined.

\section{Experimental Results}\label{sec::results}
\begin{figure}[t!]
     \centering
     \begin{subfigure}[b]{0.48\textwidth}
         \centering
        \includegraphics[width=\textwidth]{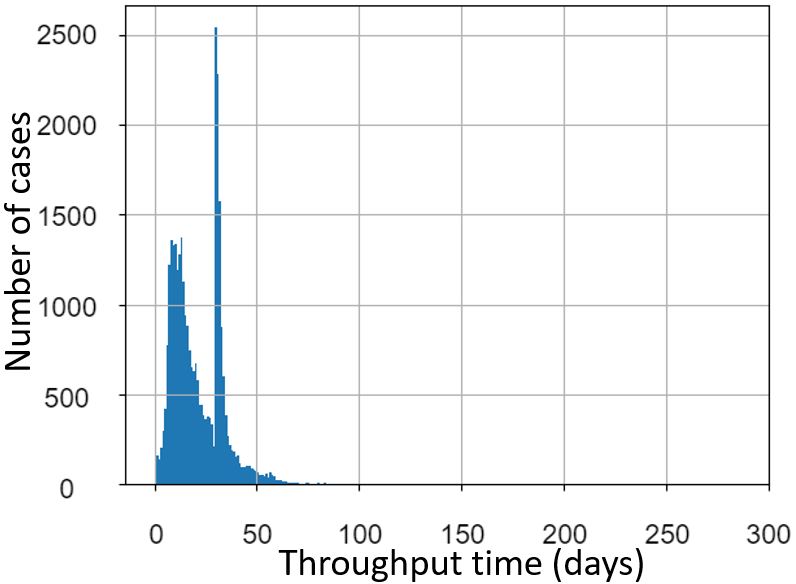}
    	\caption{Distribution of the throughput time for the BPI Challenge 2017 event log capturing the duration from the start to the end of each case.}
    	\label{fig:throughput-celonis-bpi17}
     \end{subfigure}
     \hfill
     \begin{subfigure}[b]{0.48\textwidth}
         \centering
        \includegraphics[width=\textwidth]{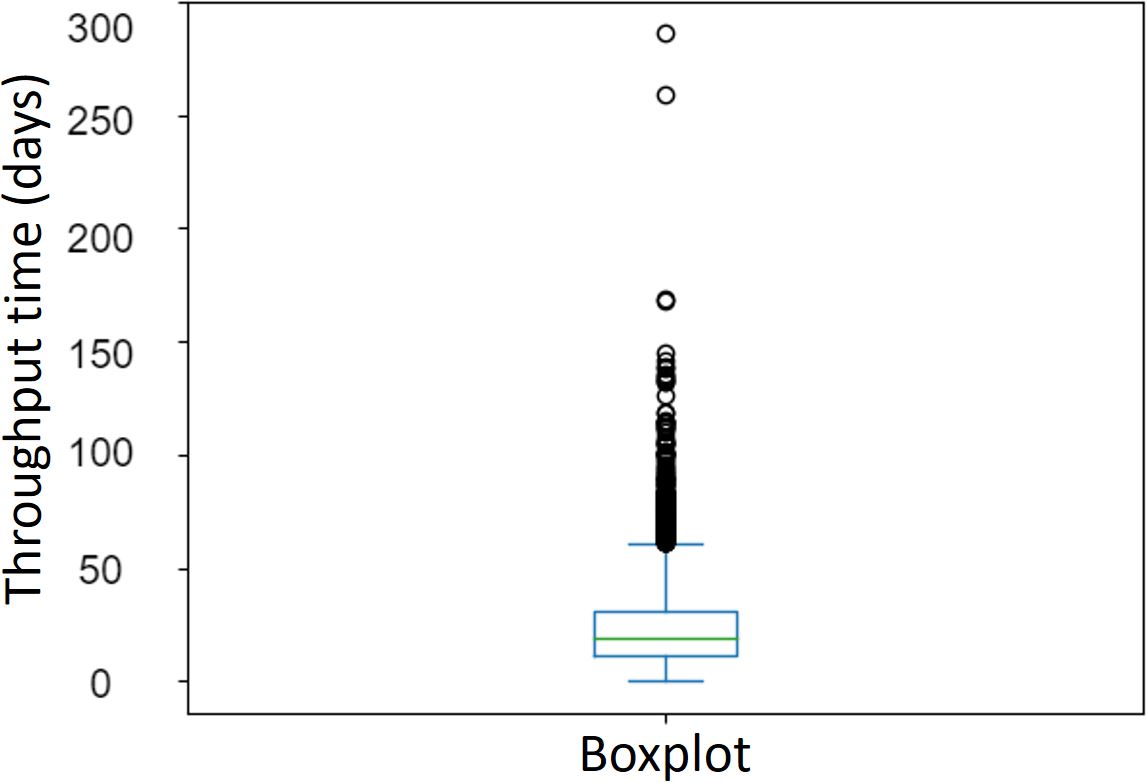}
    	\caption{Detected outliers of throughput time of cases of BPI Challenge 2017 event log using boxplot. Cases durations above 61 days are considered anomalous.}
    	\label{fig:boxplot-bpi-2017}
     \end{subfigure}
        \caption{The throughput time for the BPI Challenge 2017 event log.}
        \label{fig:case-duration-bpi17}
\end{figure}

To evaluate the proposed framework\footnote{The implemented tool is available at \url{https://github.com/ckohlschm/detecting-surprising-instances}.}, we present the result of applying it on the event log for BPI Challenge 2017 \cite{Carmona2021}. This event log  represents an application process for a personal loan or overdraft within a global financing organization taken from a Dutch financial institute. We consider throughput as the target feature. The majority of the cases in the process take between 5 and 40 days. The average duration for all cases in the event log is around 22 days. Figure \ref{fig:throughput-celonis-bpi17} shows the distribution of the throughput time. 

Boxplots are frequently used to identify performance anomalies \cite{conforti2016filtering}. %\textcolor{green}{''Data Mining: Concepts and Techniques, 3rd edition'' (https://dblp.org/rec/books/mk/HanKP2011.bib) supports this statement in general, specific to process mining there are also a few examples: ''Filtering Out Infrequent Behavior from Business Process Event Logs'' (https://dblp.org/rec/journals/tkde/ConfortiRH17.bib), ''Process Mining to Explore Variations in Endometrial Cancer Pathways from GP Referral to First Treatment'' (https://dblp.org/rec/conf/mie/KurniatiRZHHJ21.bib)} 
Thus we use boxplots as the baseline and call this approach the \emph{baseline}. The resulting boxplot is shown in Figure  \ref{fig:boxplot-bpi-2017}. Using this method, 255 cases with a throughput of more than 61 days have been considered anomalous. These are the detected anomalies without any context-awareness of the process. We also used boxplots to detect surprising situations in a vicinity cover composed of a set of 25 non-overlapping randomly selected vicinities. However, we do not present the results of this method as the detected anomalies were almost identical to the ones detected by the baseline method.

To apply our approach, we used the following case-level attributes as descriptive features: \emph{application type}, \emph{loan goal}, \emph{applicant's requested loan amount}, and the \emph{number of offers} which is a derivative attribute indicating how many times the loan application institute offered a loan to the customer. Note that in this experiment, each case is a situation.

%starting with the decision tree method to detect performance anomalies in this event log.

We apply surprising situation detection using a similarity measure, a classification method (using a decision tree), and also a clustering method (using $k$-means clustering). We call these three approaches \emph{similarity based method}, \emph{decision tree method}, and \emph{$k$-means clustering method} respectively. In all these methods, to maximize the applicability of the implemented tool and to minimize the required domain knowledge, we use the boxplot as the detector function (Definition \ref{def::ss}) to find the surprising situations in each vicinity.
\begin{figure}[t!]
	\centering
    \includegraphics[width=80mm]{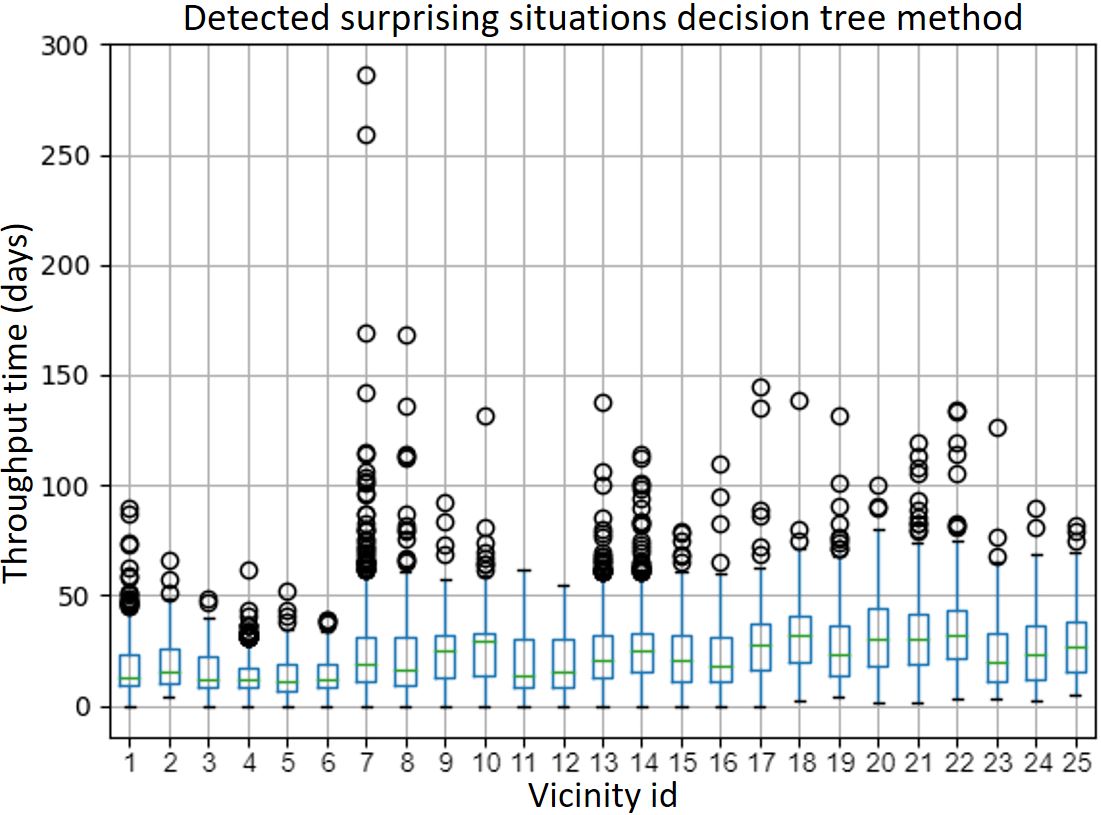}
	\caption{Detected surprising situations in each vicinity defined by the decision tree method.}
	\label{fig:boxplot-vicinity}
\end{figure}
\paragraph{Decision tree method.} For this experiment, we trained a decision (regression) tree with a maximum depth of 5 and a minimum number of instances per leaf of 100. For simplicity, we  consider the vicinities described by the leaves of the tree. Figure \ref{fig:boxplot-vicinity} shows the detected surprising situations for the leaves in the decision tree where each leaf is labeled with a number.

\begin{figure}[b!]
     \centering
     \begin{subfigure}[b]{0.48\textwidth}
         \centering
        \includegraphics[width=\textwidth]{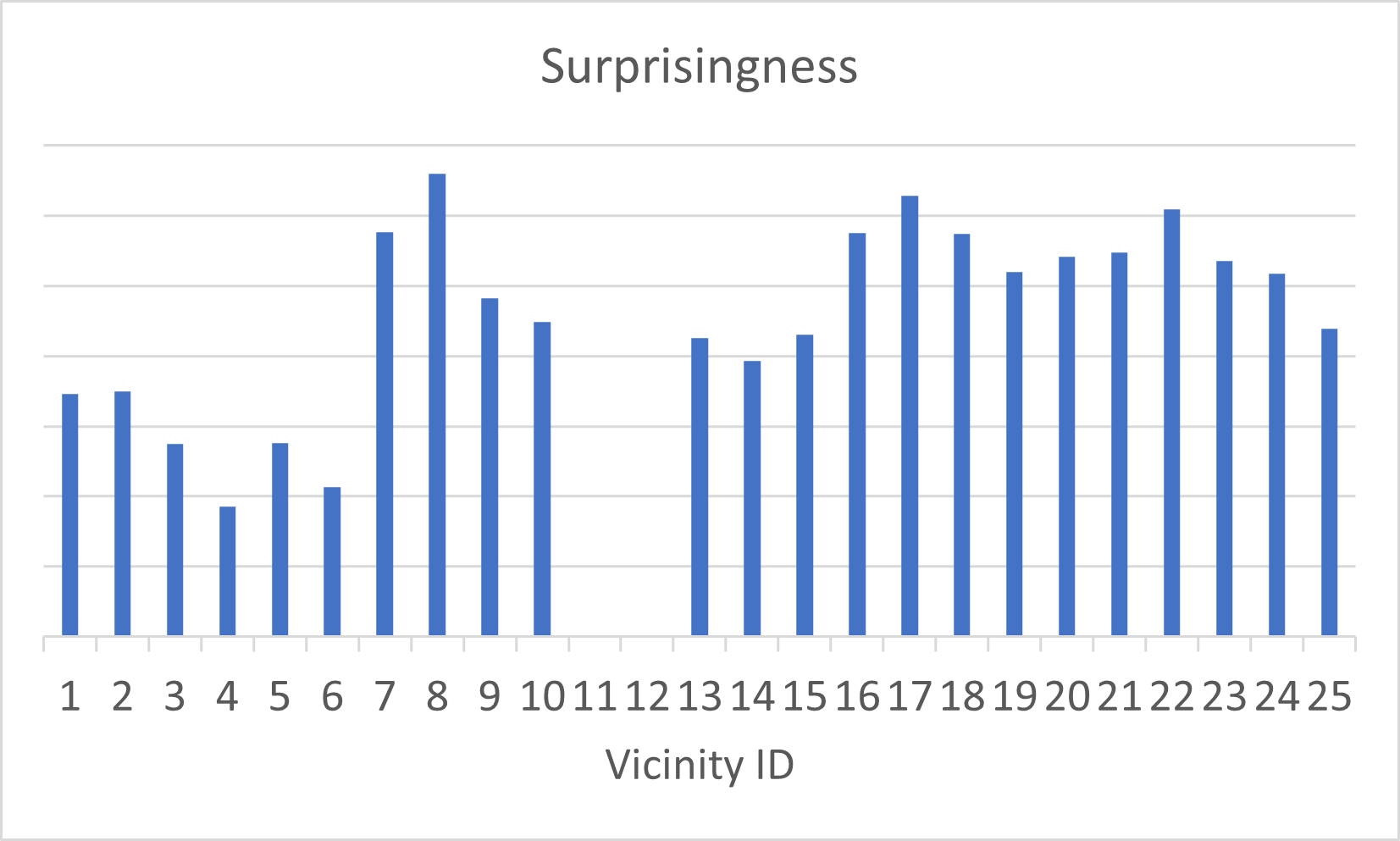}
    	%\caption{\textcolor{green}{Surprisingness measure for each vicinity detected by the decision tree method.}}
    	\label{fig:surprisingness-bpi17-decisiontree}
     \end{subfigure}
     \hfill
     \begin{subfigure}[b]{0.48\textwidth}
         \centering
        \includegraphics[width=\textwidth]{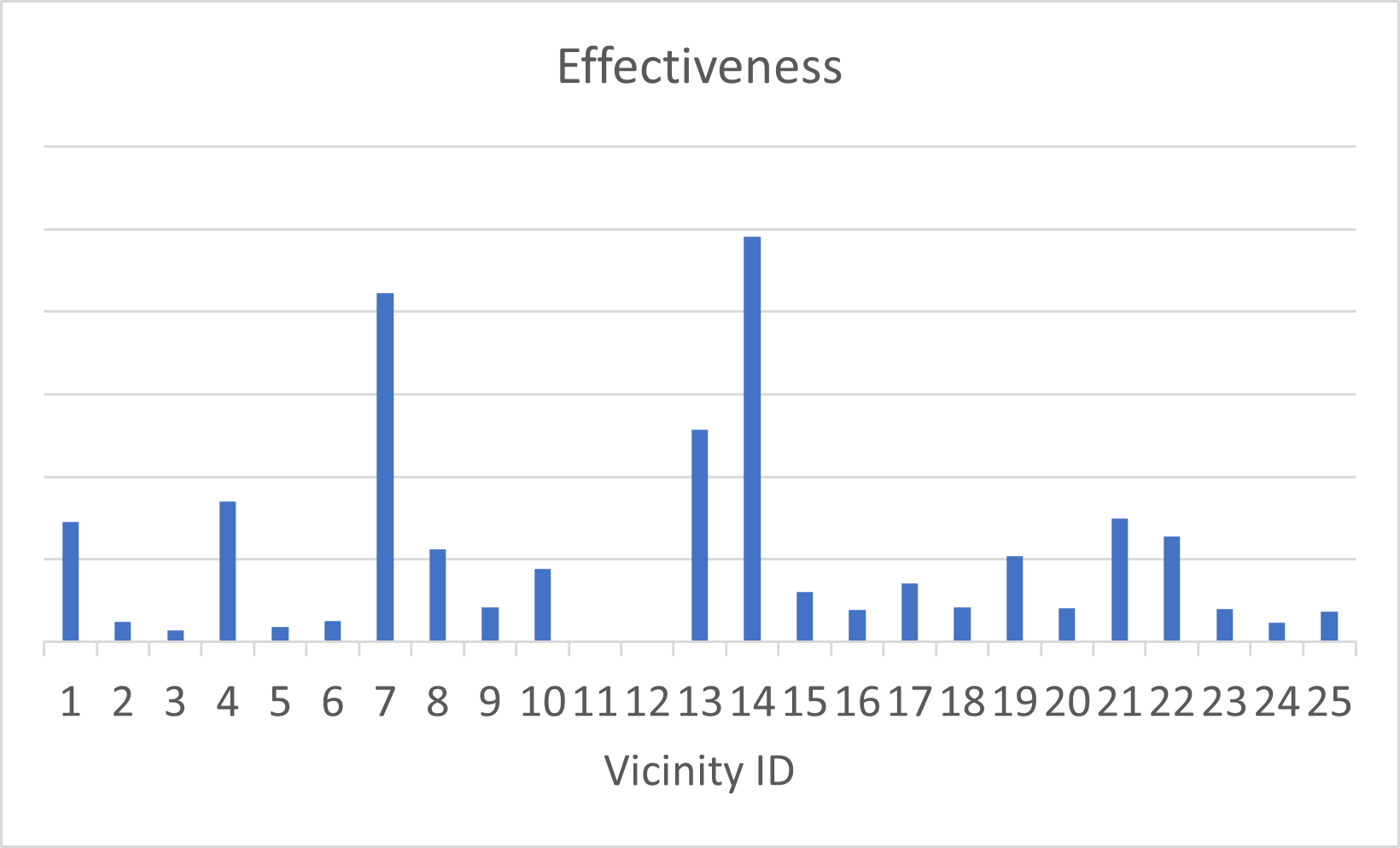}
    	%\caption{\textcolor{green}{Effectiveness measure for each vicinity detected by the decision tree method.}}
    	\label{fig:effectiveness-bpi17-decisiontree}
     \end{subfigure}
        \caption{Surprisingness and effectiveness for the surprising situations identified by the decision tree method.}
        \label{fig:surprisingness-effectiveness}
\end{figure}
%In comparison with the results of the baseline-1, we can see that 189 cases have been found surprising by both methods, 66 cases have been found just by the boxplot method, and 73 cases just by the decision tree method. 
Some of the highlights of the comparison of the results of the decision tree method and the baseline-1 are as follows:
\begin{itemize}
    % \item Case duration increases with number of offers.
    \item Application\_1839367200 (Case duration 62 days) is barely considered an outlier in the total dataset, but in its vicinity (Vicinity 4: one offer, limit raise, loan goal car, requested amount > 11.150) it is far from the average which is 14 days.
    \item Vicinity 19, where the number of offers is more than 3 and the requested amount $\leq$ 13.162 includes seven surprising situations. These situations have not been considered outliers by the baseline-1 method. One possible interpretation of this result is that high throughput is acceptable in such situations. The same applies to vicinity 20. 
%    \item Application\_1710591980 with the longest throughput (80 days) was detected as an outlier using boxplot method but using decision tree it is not detected as surprising. In vicinity 20, which this case belong to, cases are not surprising up to a duration of 81 days.
    \item Vicinity 5 (one offer, limit raise, Unknown loan goal, requested amount $\leq$ 3000) contains 3 surprising situations that are all overlooked by the baseline-1 method. The vicinity contains 338 cases with an average throughput time of 13 days which makes cases with a duration of more than 40 days surprising. The same applies to vicinities 3 and 6.
\end{itemize}
\begin{figure}[b!]
     \centering
     \begin{subfigure}[b]{0.48\textwidth}
        \centering
        \includegraphics[width=\textwidth]{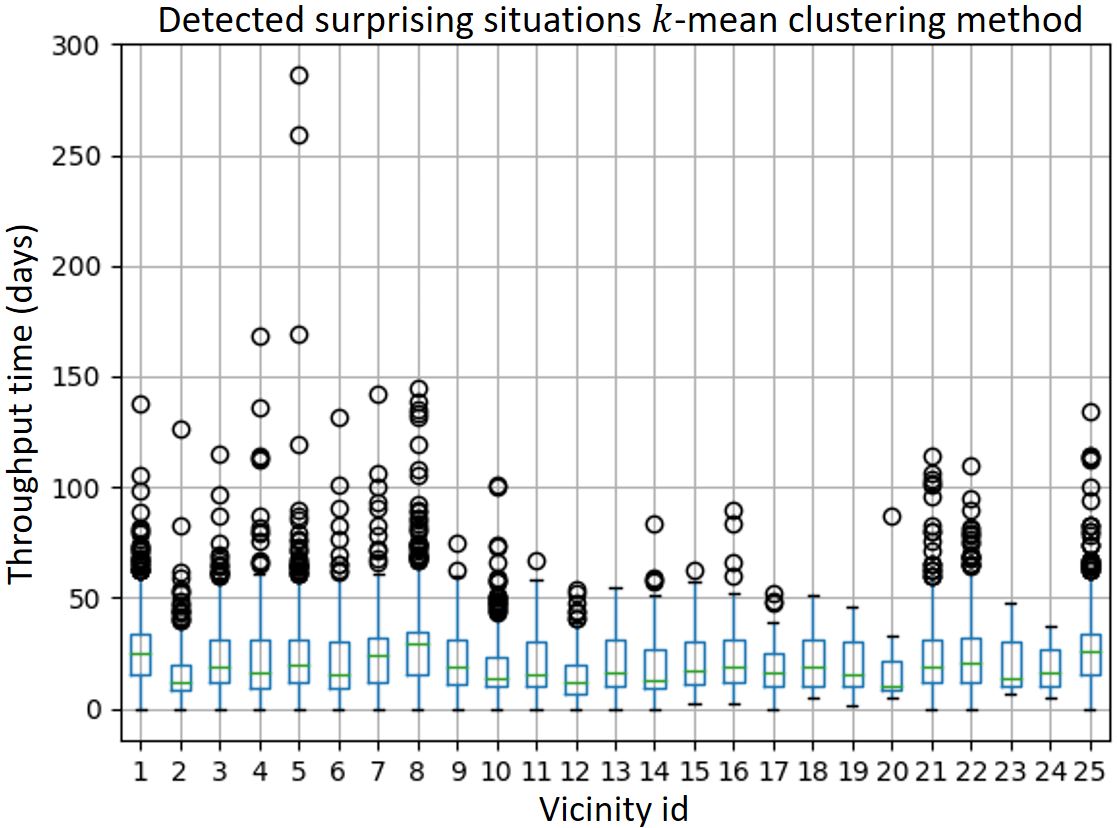}
	    \label{fig:boxplot-vicinity-kmeans}
     \end{subfigure}
     \hfill
     \begin{subfigure}[b]{0.48\textwidth}
        \centering
        \includegraphics[width=\textwidth]{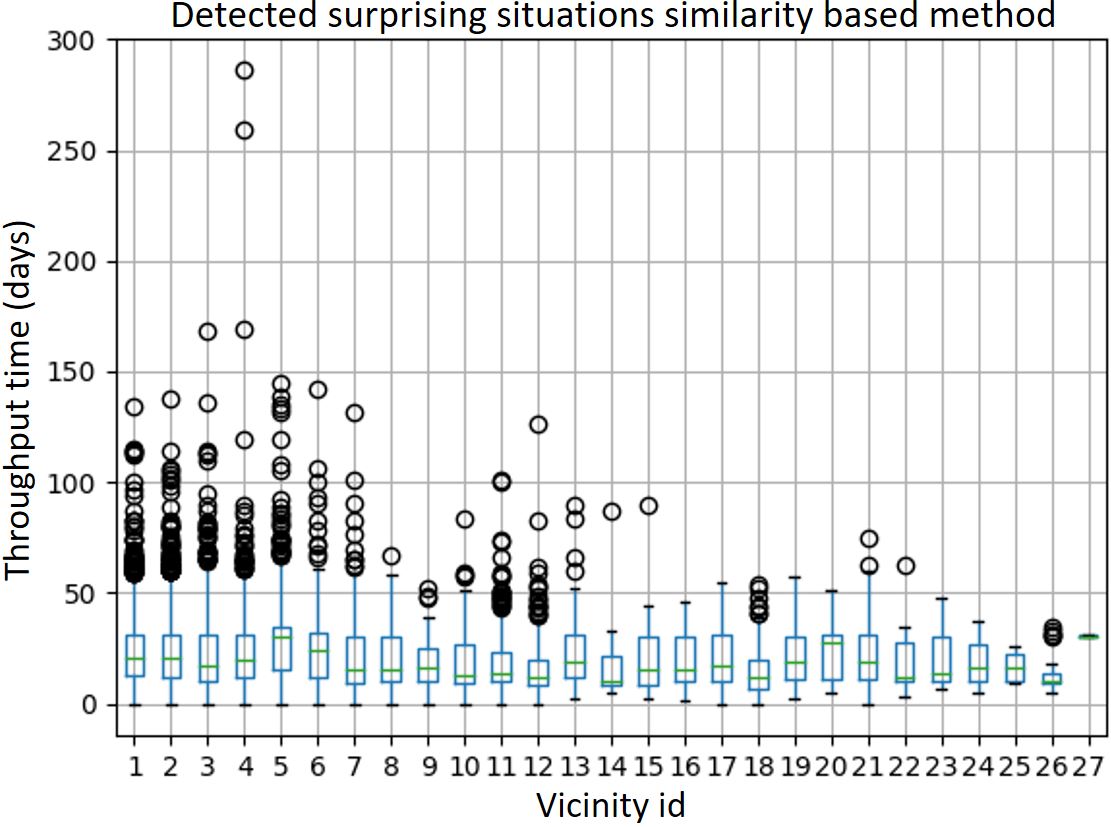}
	\label{fig:boxplot-vicinity-similaritygraph}
     \end{subfigure}
        \caption{Detected surprising situations for the $k$-means clustering and similarity based method.}
        \label{fig:kmeanAndSimilarity}
\end{figure}

Figure \ref{fig:surprisingness-effectiveness} shows the surprisingness (on the left) and effectiveness (on the right) of the sets of surprising situations detected by the decision tree method. The set of surprising situations in vicinity 8 has the highest surprisingness. This vicinity includes 6136 situations, where 13 are surprising with an average throughput of 96 days, whereas the other situations in the vicinity have an average of 19 days. The set of surprising situations in vicinity 14 has the highest effectiveness. Removing the problem that causes the delay in these surprising situations would reduce the average throughput time for similar cases by more than one day.

%result in 4.06\% increase in the performance of the process in similar cases.

% Average duration with surprising situations (total process): 21.899 days
% Average duration without surprising situations (total process): 21.829 days
% Result: around 0.32\% decrease of case duration which is around 1 hour and 40 minutes that would be saved on average in the overall process if the problem that caused the delay would be removed

% Average duration with surprising situations (vicinity): 26.33 days
% Average duration without surprising situations (vicinity): 25.26 days
% Result: around 4.06\% decrease of case duration in similar cases if the problem that caused the delay would be removed

\paragraph{$k$-means clustering method.}
In this approach, we used $k$-means clustering to identify vicinities. For $k$ we use the value 25, which is the number of the leaves of the decision tree model in the previous part of the experiment. This method results in detecting a total of 280 surprising situations. The plot on the left side of Figure \ref{fig:kmeanAndSimilarity} shows the surprising situations detected in each vicinity.

\paragraph{Similarity based method. }We run the similarity based approach where the distance measure is the Euclidean distance of normalized descriptive features (using min-max method). Then, we use 1.4 as the threshold to generate a graph. To find the vicinities, we used the  \emph{Louvain community detection method} \cite{Blondel_2008} on this graph. The plot on the right side of Figure \ref{fig:kmeanAndSimilarity} shows the surprising situations detected in each vicinity. %consists of two phases. The first phase is the modularity optimization phase. The algorithm first detects small communities by optimizing modularity locally on all nodes. Therefore, it assigns every node to be in its own community. Then, for each node the algorithm tries to find the maximum modularity gain by moving the node to all of its neighbor communities. If no positive gain is achieved, the node remains in its original community.We use  which leads to a reasonable amount of communities. For the feature vectors, we normalize the columns using 0,1 normalization. 

\begin{figure}[t!]
	\centering
    \includegraphics[width=70mm]{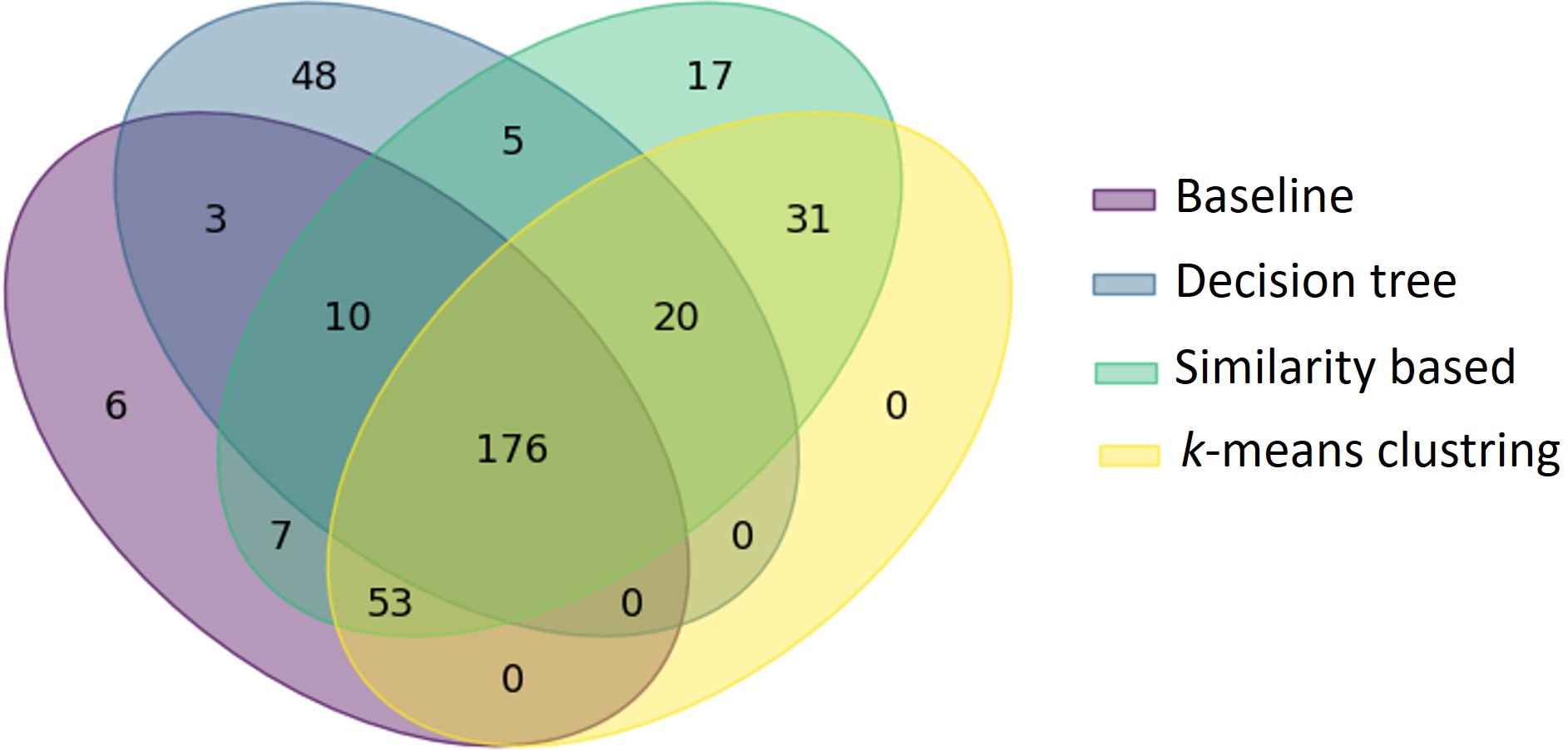}
	\caption{Venn Diagram showing the intersection of detected surprising situations using the different methods.}
	\label{fig:venn-diagram-comparison}
\end{figure}

It is worth noting that the set of surprising situations detected by different methods was not exactly the same. Figure \ref{fig:venn-diagram-comparison} shows that all the methods agree on 176 detected surprising situations and for all other situations at least one method does not select it.

%\begin{figure}[h]
%	\centering
%    \includegraphics[width=\textwidth]{AnomaliesForVicinity.PNG}
%	\caption{Identified anomalies for each set of similar cases}
%	\label{fig:num-anomalies-vicinity}
%\end{figure}

\section{Conclusion}\label{sec::conclusion}

Finding the process enhancement areas is a fundamental prerequisite for any process enhancement procedure that highly affects its outcome. It is usually assumed that these process areas are known in advance or can be detected easily. However, utilizing simple methods have the danger of overlooking some of the opportunities for process enhancement or targeting the wrong ones. In this paper, we formulate the process of finding process enhancement areas as a method for finding surprising situations; i.e., detecting those situations where the process behavior is significantly different from similar situations.

We have implemented the proposed framework with different methods and evaluated it using real event logs. The experiment shows that the detected surprising (anomalous) situations are overlapping but not identical to the ones of the baseline, which is currently a common method for finding anomalies. It shows that to find the best result, it is best to use our framework complementary to the existing methods; i.e., using both context-sensitive and non-context-sensitive methods for finding the process enhancement areas. 

%In future work, we aim to more deeply analyze the qualitative difference between the outcomes of the proposed approaches to find surprising situations. Moreover, extending this method to an almost automated approach for root cause analysis is another direction for the future work.
\section*{Acknowledgment}
We thank the Alexander von Humboldt (AvH) Stiftung for supporting our research.
\bibliographystyle{abbrv}
\bibliography{biblio}

\end{document}